\documentclass[10pt,twocolumn,letterpaper]{article}

\usepackage{cvpr}              % To produce the CAMERA-READY version
\usepackage{stfloats}

\definecolor{cvprblue}{rgb}{0.21,0.49,0.74}
\usepackage[pagebackref,breaklinks,colorlinks,allcolors=cvprblue]{hyperref}

\def\paperID{1971} % *** Enter the Paper ID here
\def\confName{CVPR}
\def\confYear{2025}

\title{Advantages of Neural Population Coding for Deep Learning}

\author{Heiko Hoffmann\\
Magimine, LLC;
Simi Valley, CA 93065, USA;
{\tt heiko@magimine.com}
}

\begin{document}
\maketitle

\begin{abstract}
Scalar variables, e.g., the orientation of a shape in an image, are commonly predicted using a single output neuron in a neural network. In contrast, the mammalian cortex represents variables with a population of neurons. In this population code, each neuron is most active at its preferred value and shows partial activity for other values. Here, we investigate the benefit of using a population code for the output layer of a neural network. We compare population codes against single-neuron outputs and one-hot vectors. First, we show theoretically and in experiments with synthetic data that population codes improve robustness to input noise in networks of stacked linear layers. Second, we demonstrate the benefit of using population codes to encode ambiguous outputs, such as the pose of symmetric objects. Using the T-LESS dataset of feature-less real-world objects, we show that population codes improve the accuracy of predicting 3D object orientation from image input.
\end{abstract}

\section{Introduction}

In the mammalian cortex, many variables, e.g., object orientation \cite{Hubel1959} and movement direction \cite{georgopoulos1986}, have been found to be encoded by populations of neurons. In a population code, each neuron responds maximally to its preferred value of an encoded variable and partially to other values. Activation levels are shaped by tuning curves such as Gaussian or cosine functions of the distance between encoded and preferred value. The activity of the group of neurons resembles a probability distribution of the encoded variable \cite{Pouget2000}.

A lot of computational neuroscience work, particularly the earlier work, has focused on decoding the information represented by a population code \cite{seung1993,salinas1994,zemel1998,baldi1988}. But the brain does not need to decode population codes: information is processed from population code to population code throughout the cortex \cite{desimone1985, pouget1998, yang2004, hoffmann2024}. 

Commonly used artificial neural networks, such as convolutional neural networks (CNNs) and multi-layer perceptrons (MLPs), encode information also by groups of neurons, particularly in their intermediate layers \cite{hinton2002}. However, in the output layer, information is typically mapped onto a different representation. For classification tasks, outputs are commonly represented as one-hot vectors, using one neuron for each classification label. For prediction tasks, output neurons typically correspond to the variables of interest, such as the position and orientation of an object in an image.  

Here, we investigate the benefit of mapping onto population codes for prediction tasks. Other prior work already pointed out the benefits of neural population codes: such coding has been shown to improve linear separability of temporal information \cite{pan2019}.  In addition, population codes can be used in the output layer for cleaning up noisy signals nearly as optimal as the maximum likelihood estimate \cite{pouget1998}. Different here, we demonstrate that replacing prediction target variables with population codes improves noise robustness and accuracy, which we show using theoretical analysis and experiments with synthetic and real-world data.

In the remainder of this article, we first derive theoretically the noise robustness for a single-layer linear network and compare single-variable, population-code, and one-hot vector outputs. Here, we include the one-hot vector for the prediction task due to its structural similarity to the population code:  it uses the same group of neurons, only the target activations differ, which are binary for one hot and continuous for the population code. Secondly, we compute the noise robustness using deeper MLPs in simulation. Here, we found that population codes lead to sparser information flow through the network compared to one-hot vectors. Finally, we use an image-to-pose prediction task to demonstrate that population codes can handle ambiguous poses from symmetric objects while improving accuracy.

This article makes the following three main contributions:

\begin{enumerate}
\item Introducing population codes as output layers of CNNs and MLPs for prediction tasks and demonstrate their benefit,
\item Analyzing theoretically the robustness to noise of single-layer networks, and
\item Discovering that training networks with population-code outputs results in sparser information flows.
\end{enumerate}

\section{Theory}

For simplicity and theoretical tractability, we investigate single-layer linear networks. First, we analyze noise robustness for networks with a single output variable, followed by networks with one-hot vector and population code outputs.

\subsection{Single-Variable Output}

Let ${\bf x}$ be the input vector and $y$ be the single-variable output, 
\begin{equation}
y = {\bf w}^T {\bf x} + b\,,
\end{equation} 
where {\bf w} is the weight vector and $b$ the bias. 

As training data, we assume that ${\bf x}$ is a one-dimensional image of size $n$, where all pixel values are zero except one that equals 1, i.e., the image contains a 1-pixel shape in arbitrary location. The target output $y$ is the location of this pixel, $y = i / n$, where $i$ is the index of the input pixel that equals 1. While simplified, this setting is also relevant to a wide range of network architectures, where at the output, a linear layer maps an encoding resembling a population code onto a single variable. Here, we assume that the training data contains all possible values of $i$.

We train this network with a squared error loss, resulting in the following weight updates,
\begin{eqnarray}
w_i^{t+1} &=& w_i^t + \eta \left( \hat{y} - y \right) x_i \\
b^{t+1} &=& b^t  + \eta \left( \hat{y} - y \right)\,,
\end{eqnarray}
where $\hat{y}$ is the target value, and $\eta$ is the learning rate. For our specific training input, $x_j = \delta_{ij}$ (Kronecker delta) for the target $\hat{y} = i/n$, the weight and bias updates simplify to
\begin{eqnarray}
w_i^{t+1} &=& w_i^t + \eta \left( i/n - w_i^t - b^t\right)\label{eq:w_update}\\
b^{t+1} &=& b^t + \eta \left( i/n - w_i^t - b^t\right)\label{eq:b_update} \,.
\end{eqnarray}
For the weights and bias to converge, the following equality must hold,
\begin{equation}\label{eq:learned_w}
w_i = \frac{i}{n} - b\,.
\end{equation}
Since this equality can be fulfilled for any $b$ value, the weights are defined only up to a constant bias term that shifts all weight values. The actual value of $b$ will depend on network initialization.

Next, we probe the network's robustness to noise in the input. For this test, we perturb the input ${\bf x}$ by ${\bf d}$, so that ${\bf x} + {\bf d}$ is the new input. Moreover, let $d_j = a \delta_{jk}$ for a given perturbed neuron $k$, and we use the same ${\bf x}$ and target, $i/n$, as above. As a result, the output equals,
\begin{equation}
y =  \frac{i}{n} + a w_k\,,
\end{equation}
with error $a w_k$. 

Assuming an error tolerance of $1.5/n$ (essentially, we tolerate a position error of one pixel but not two), we want to compute the rate of network failures. To compute the failure rate, we approximate that the trained weights will be in the range $-0.5$ to $0.5$ since the weights are commonly initialized to be uniformly distributed with zero mean, and after training, their range is about 1. According to (\ref{eq:learned_w}), the trained weights are uniformly distributed. Moreover, we assume that the target and perturbation indices are drawn with uniform probability from  $\{0, 1, \dots, n-1\}$. For the trials that fall within the error tolerance, we have 
\begin{equation}\label{eq:w_condition}
|w_k| < \frac{1.5}{a n}\,.
\end{equation}
The ratio of trials, $r$, within the error tolerance is the fraction of $w$ values fulfilling (\ref{eq:w_condition}) compared to the total weight range after training,
\begin{equation}
r = \min(1, \frac{3}{a n})\,.
\end{equation}
Here, we approximated the discretely distributed values with continuous uniform distributions. As a result, the failure rate is
\begin{equation}
r_F = \max(0, 1 - \frac{3}{a n})\,.
\end{equation}
For example, for $n=20$, even for a relatively small perturbation of $a = 0.2$, we expect a failure rate of 25\%. At $a=0.5$, the failure rate increases to 70\%.

\subsection{Population-Code Output}

In comparison, we investigate a linear network with population-code output, ${\bf y}$, instead of a single variable, 
\begin{equation}\label{eq:model_pop}
{\bf y} = {\bf W} {\bf x} + {\bf b}\,,
\end{equation}
where ${\bf W}$ is the weight matrix. For squared error loss, the weight update rule becomes
\vspace{-1mm}
\begin{equation}
w_{oi}^{t+1} = w_{oi}^t + \eta \left(\hat{y}_o - \sum_j w_{oj}^t x_j - b_o\right) x_i\,.
\end{equation}

For our theoretical analysis, we first use a simplified population code that resembles a one-hot vector, i.e., only one neuron, $y_o$, with preferred value $o/n$ gets active, and all other neuronal activations are zero. Later, we will show the difference between population code and one-hot vector. 

For training, the target vector matches the input; so, the network has to learn the identity function (our arguments also hold for learning any permutation of the indices, e.g., a shift in pixel position, because the assignment of neurons to indices is arbitrary). 

Using the training input, $x_j = \delta_{ij}$ and target $\hat y_o = \delta_{oi}$ with input value 1 at index $i$, the weight update simplifies to 
\begin{equation}\label{eq:w_update_pop}
w_{oi}^{t+1} = w_{oi}^t + \eta \left(\delta_{oi} - w_{oi}^t - b_o \right)\,.
\end{equation}
This update converges to 
\begin{equation}\label{eq:learned_w_pop}
w_{oi} = \delta_{oi} - b_o\,.
\end{equation}

For computing the prediction error as above, we need to decode the population code. Here, we use a simple decoding rule, namely, we read out the preferred value of the neuron that is the most active, i.e., 
\begin{equation}
v = \frac{\underset{j}{\operatorname*{arg\,max}} \, y_j}{n}
\end{equation}

As above, we perturb the input by $d_j = a \delta_{jk}$. Using  (\ref{eq:model_pop}) and (\ref{eq:learned_w_pop}), the perturbed input produces the output
\begin{equation}\label{eq:pop_output}
y_o = \delta_{oi} + a \delta_{ok} - a b_o\,.
\end{equation}
For a failure to occur, a necessary condition is $y_k  > y_i$. Using (\ref{eq:pop_output}), this condition is equivalent to
\vspace{-1mm}
\begin{equation}
a > \frac{1}{1 - b_k + b_i} 
\end{equation}
if $1 - b_k + b_i > 0$. Since the bias is commonly initialized to zero and due to the symmetry of our problem with respect to the indices, we expect only a minor difference between bias values. Therefore, $a$ has to be near 1 for a failure to occur, which makes the population code more robust to noise compared to the single variable output.  

For an actual population code, the activation values follow a tuning curve rather than being binary. So, the allowable threshold for $a$ will be slightly smaller, as shown in the following. For Gaussian tuning, a failure occurs if the maximum of the sum of the Gaussians from the real signal and perturbation is shifted by at least $2/n$ compared to the target value (the preferred values are discrete in steps of $1/n$). This condition necessitates that the activation at target is smaller than this maximum, 
\vspace{-1mm}
\begin{eqnarray}\label{eq:max_Gauss_sum}
1 + a \exp\left(-\frac{(4/n)^2}{2\sigma^2}\right) < \nonumber\\ \max_{x \ge 2/n}  \left( \exp\left(-\frac{x^2}{2\sigma^2}\right) +  a \exp\left(-\frac{(4/n - x)^2}{2\sigma^2}\right) \right)\,,
\end{eqnarray}
where $\sigma$ is the tuning width, and $4$ is the worst-case difference between $i$ and $k$ that could result in errors above the tolerance of $1.5/n$ for $a<1$. The right-hand side is maximized at $x = 2/n$ because the maximum cannot be at $x>2/n$ if $a<1$ due to the higher amplitude of the Gaussian at target ($x = 0$). 

For $x=2/n$, the inequality (\ref{eq:max_Gauss_sum}) becomes
\begin{equation}
1 + a \exp\left(-\frac{8}{n^2\sigma^2}\right) <  \left(1 + a\right) \exp\left(-\frac{2}{n^2\sigma^2}\right).
\end{equation}
Solving for $a$ results in 
\begin{equation}
a > \frac{1 - \exp\left(-\frac{2}{n^2\sigma^2}\right)}{\exp\left(-\frac{2}{n^2\sigma^2}\right) - \exp\left(-\frac{8}{n^2\sigma^2}\right)}.
\end{equation}
For example, for $\sigma = 0.1$ and $n = 20$, the necessary condition for failure is $a > 0.835$, which is still much better compared to the single-variable case. In comparison, for $a = 0.835$, the failure rate for the single-variable output is about 82\%.

For the single-layer network, the one-hot vector is more robust than the population code, but as we will see for deeper networks, this advantage disappears, and the population-code network becomes the most robust.

\section{Experiments With Synthetic Data}

In numerical experiments using the same training data as for the theory, we compare the three approaches on deeper networks.

\subsection{Methods}

We used various numbers of stacked linear layers of size $n = 20$, except for the output layer for the single-variable approach, which consisted of only one neuron. In the architecture, each hidden layer was followed by a leaky ReLU function. For the population-code approach, a sigmoid function was applied to the output, and we used Gaussian tuning curves with a width of $\sigma = 0.1$.

\begin{figure*}[t]
\centering
Robustness Results Without Data Augmentation\\
\includegraphics[width=0.49\textwidth]{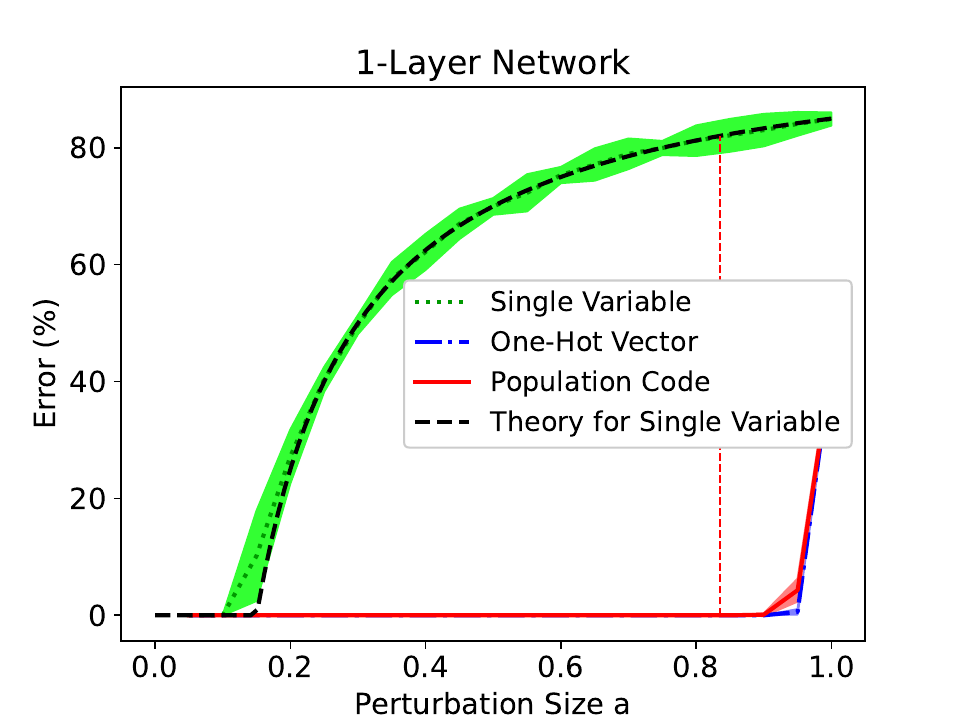}\hspace{1mm}
\includegraphics[width=0.49\textwidth]{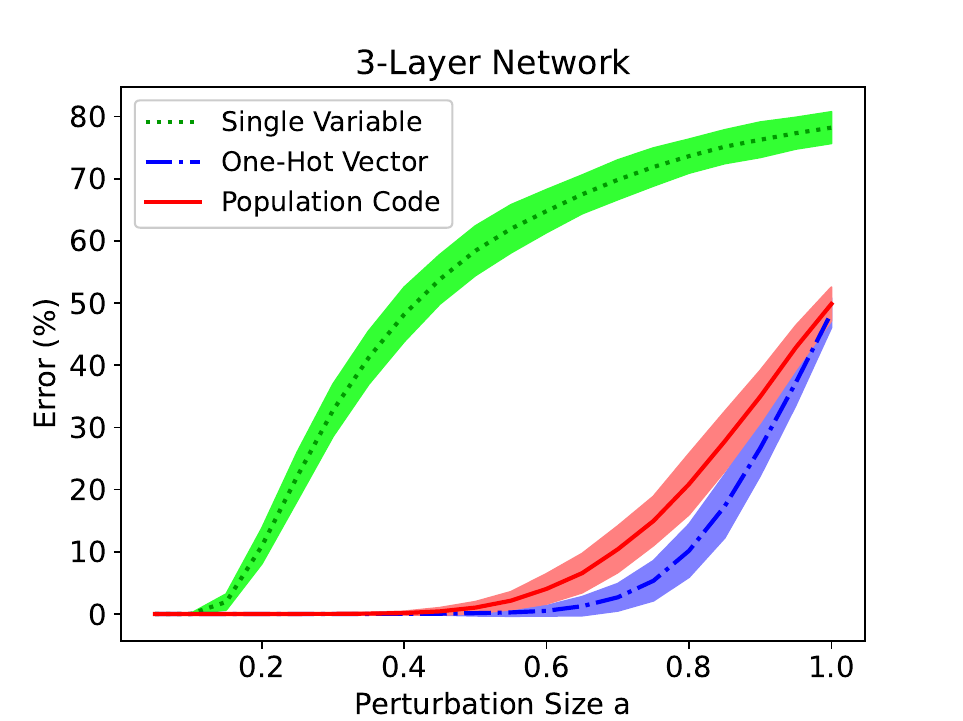}\\
\includegraphics[width=0.49\textwidth]{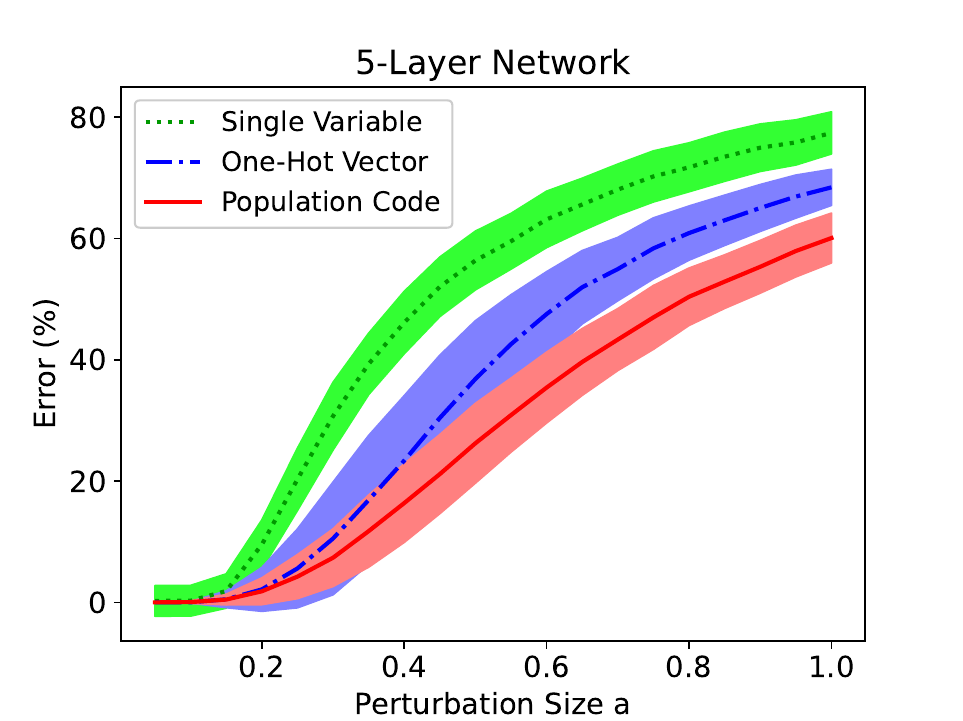}\hspace{1mm}
\includegraphics[width=0.49\textwidth]{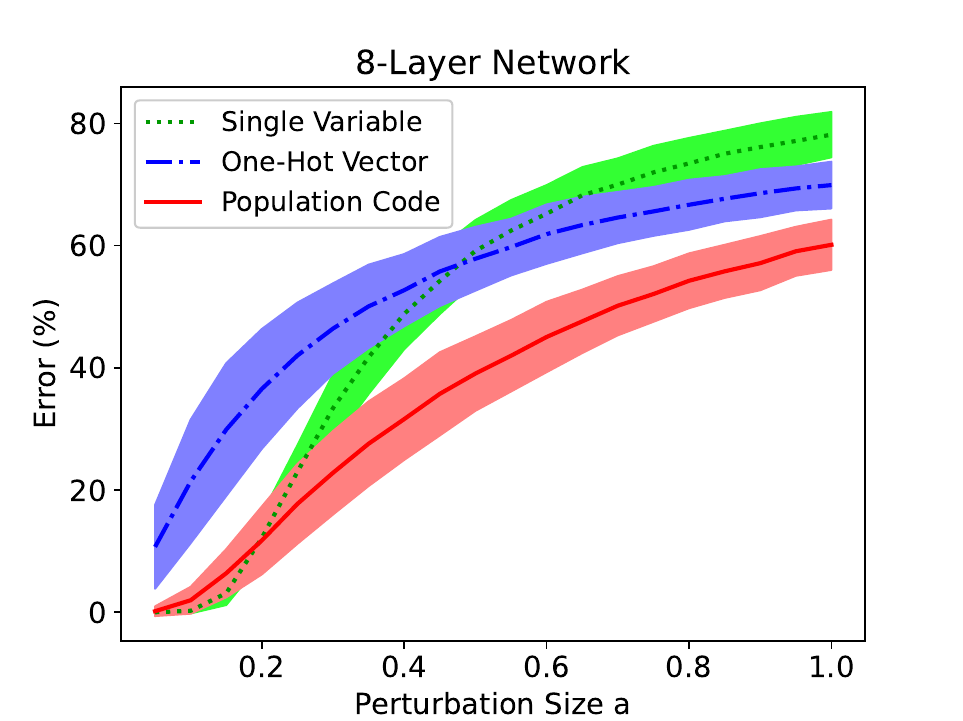}
\vspace{-2mm}
\caption{Robustness to input perturbations comparing single-variable, one-hot vector, and population-code outputs. Here, the networks were trained without input noise. The dashed thin red line in the first panel shows the theoretical bound of $a$ below which the population-code has zero errors. Experimental results show means $\pm$ SD, as indicated by shaded areas (100 simulation runs).}\label{fig:robust}
\end{figure*}

We used the mean squared error (MSE) as the loss function for single-variable and population code models, and cross-entropy loss for one-hot encoding, which is the commonly used loss function for classification tasks. The networks were implemented and trained using the PyTorch framework. For training, we used the Adam optimizer with a learning rate of $\eta = 0.005$ and $5,000$ epochs (sufficient for convergence). In each simulation run, the networks were initialized (default initialization), trained from scratch, and tested. For each method and network size, we computed 100 simulation runs. 

In each test run, for each clean input vector ($n$ different ones), we presented $1,000$ perturbations at a random input neuron (uniformly chosen) and with random amplitude $a$, uniformly chosen from the set $\{0.05, 0.1, 0.15,\dots, 1\}$. For each amplitude, we computed the failure rate across all inputs and perturbation locations and then averaged rates across all simulation runs.

\subsection{Results on Noise Robustness}

Figure \ref{fig:robust} shows the results for networks with 1, 3, 5, and 8 layers. The 1-layer network mirrored the theoretical assumptions, and the results are in good agreement with the theory. There was a small discrepancy between theory and simulation for the single-variable method for $a = 0.15$. The mismatch can be explained by noting that the weights were not exactly symmetrically distributed around zero; instead, the range was, on average,  $[-0.375; 0.575]$. This bias towards positive values can in turn be explained by the weight and bias update rules, (\ref{eq:w_update}) and (\ref{eq:b_update}), which initially push both average weight and bias from zero to positive values.

The population-code method had the best noise robustness for deeper networks with 5 and 8 layers. For shallow networks with 1 and 3 layers, the one-hot method was slightly better, but in that case, the population code was still much better than the single-variable method.

\begin{figure*}[t]
\centering
Robustness Results With Data Augmentation\\
\includegraphics[width=0.49\textwidth]{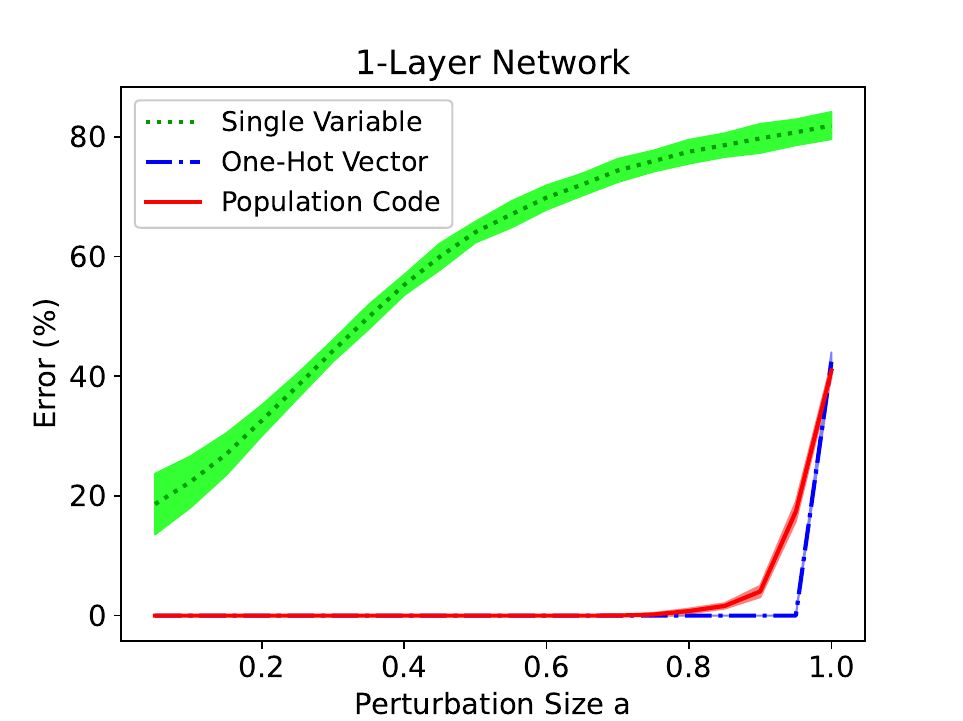}\hspace{1mm}
\includegraphics[width=0.49\textwidth]{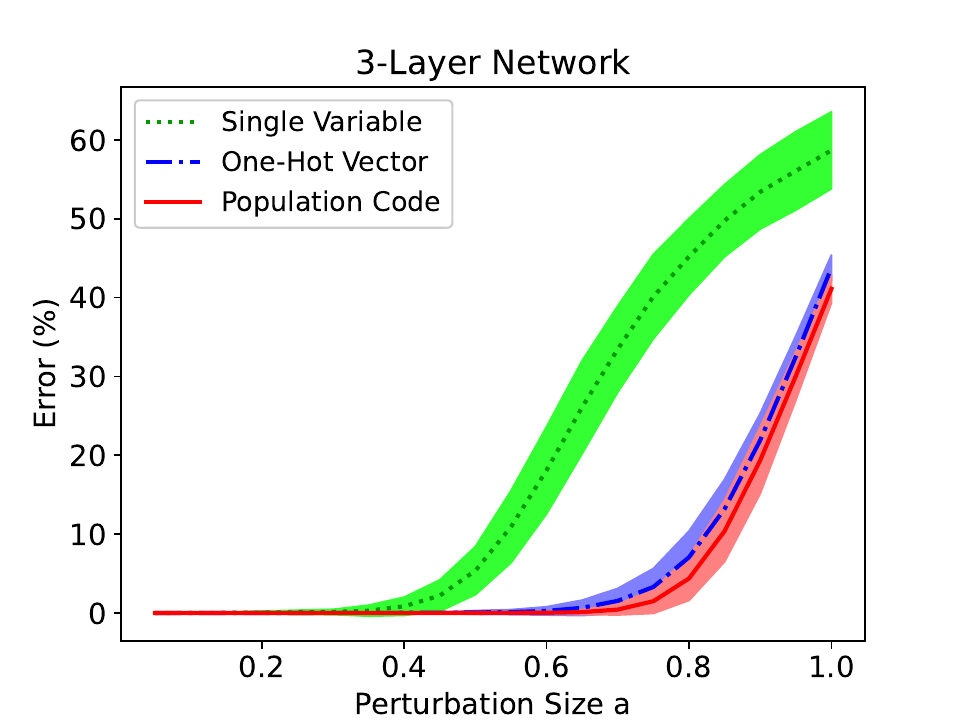}\\
\includegraphics[width=0.49\textwidth]{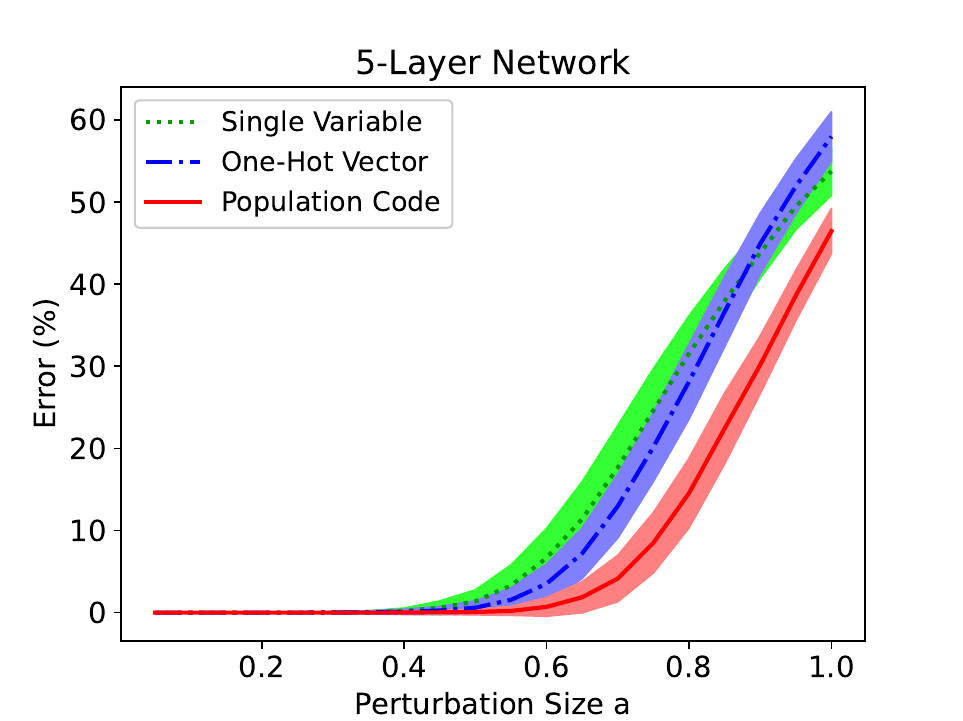}\hspace{1mm}
\includegraphics[width=0.49\textwidth]{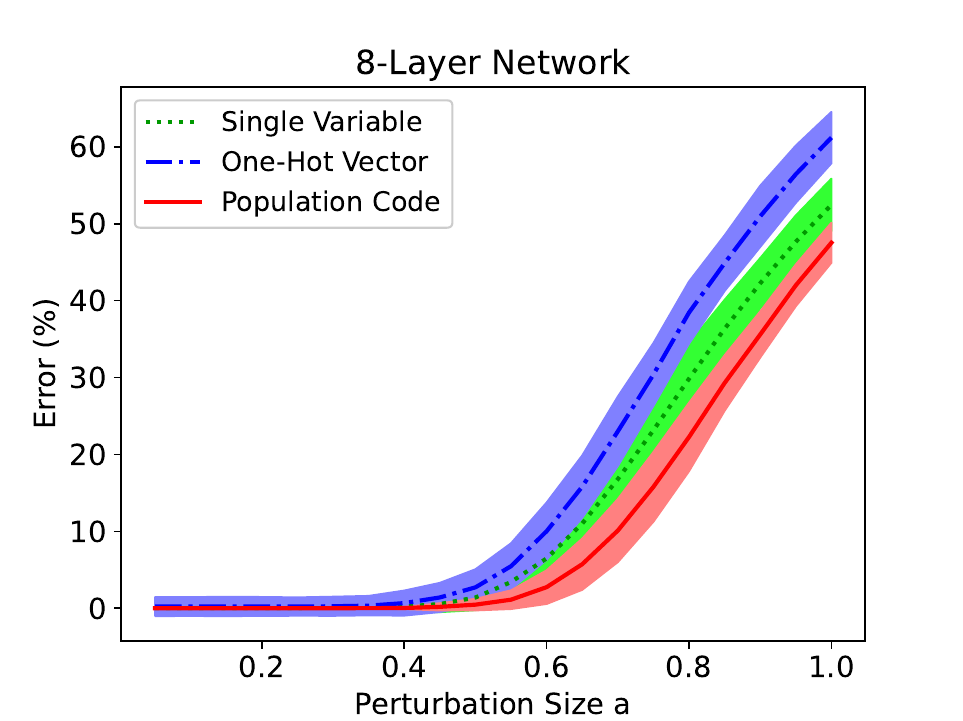}
\vspace{-2mm}
\caption{Robustness to input perturbations comparing single-variable, one-hot vector, and population-code outputs. Here, the networks were trained by adding Gaussian input noise. Experimental results show means $\pm$ SD, as indicated by shaded areas (100 simulation runs).}\label{fig:robust_augmentation}
\end{figure*}

We ran an additional experiment to test the impact of data augmentation on the results (Fig. \ref{fig:robust_augmentation}). Here, for training, we added a small Gaussian noise, $\mathcal{N}(0, 0.1^2)$, to each input pixel value. The data augmentation improved noise robustness for networks with more than one layer, but it did not change qualitatively the comparison between methods. The population code was still overall the best choice. 

\begin{figure*}[b]
\centering
\includegraphics[width=0.49\textwidth]{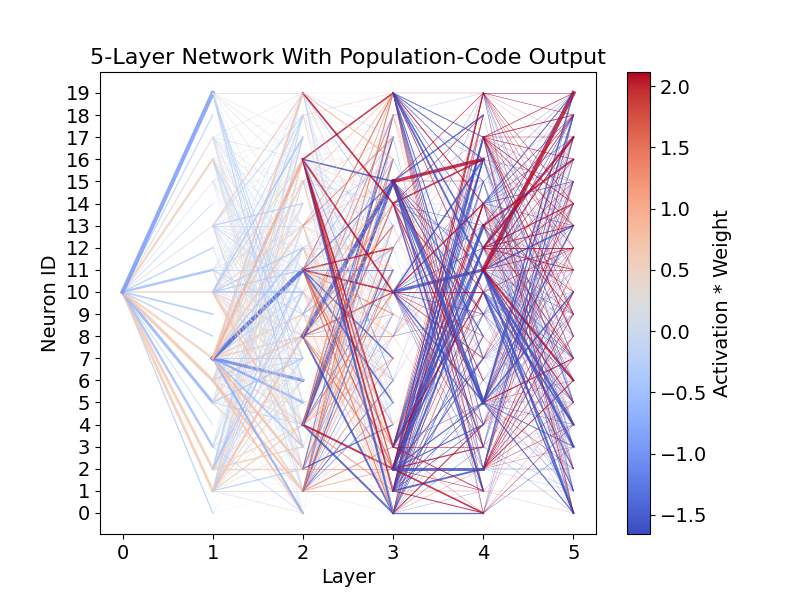}\hspace{1mm}
\includegraphics[width=0.49\textwidth]{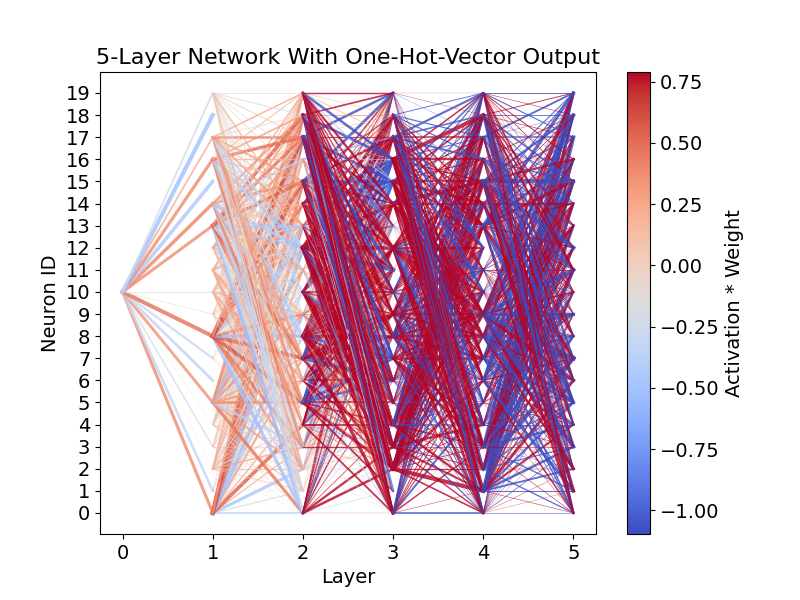}
\vspace{-2mm}
\caption{Population codes lead to sparser information flow compared to one-hot vectors. The flow is the product of the neural activity times the weight of the outgoing connection. Here, the network's input was 1 for neuron with ID 10 and zero otherwise.}\label{fig:sparse_example}
\end{figure*}

\subsection{Population Code vs One-Hot Vector}

In the following, we will investigate why the population code performed better than one hot for noise robustness. Both population code and one hot compute a sigmoid function on the output of the last linear layer; the cross-entropy loss includes a softmax on the logits. However, the difference is that for one hot, the target values are binary, while the population code has continuous values between 0 and 1. When 0 is the target for a sigmoid function, due to its fast convergence to 0, the logits can be arbitrarily large. Beyond a certain size, the loss function is insensitive to their value.

\begin{figure}[h]
\centering
\includegraphics[width=0.49\textwidth]{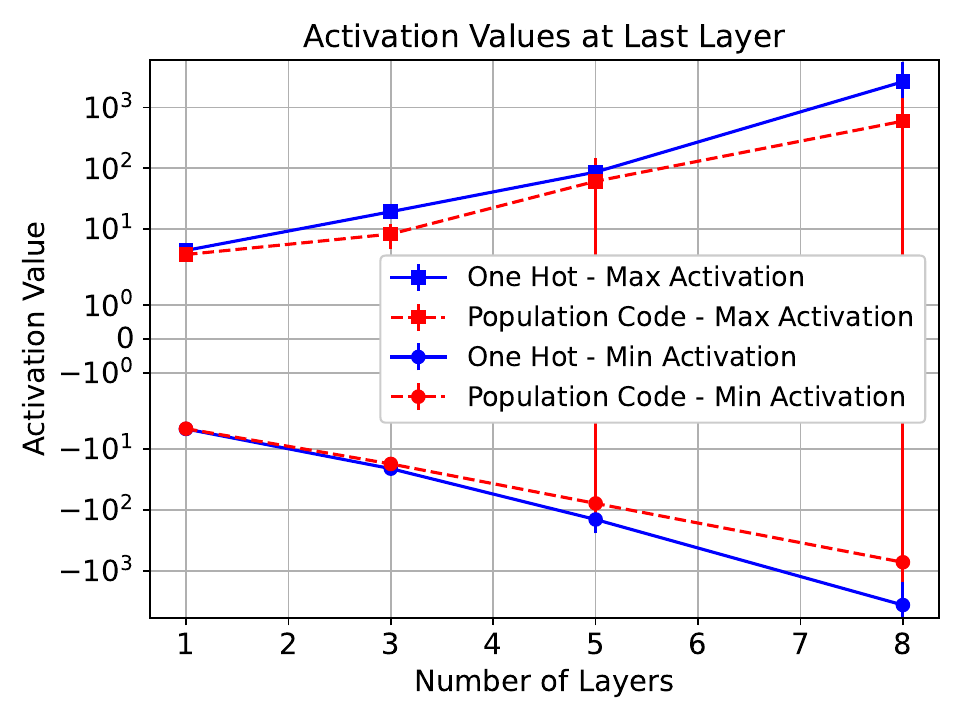}
\vspace{-5mm}
\caption{Activation values after the last linear layer were more extreme for the one-hot approach compared to the population code. Values are mean $\pm$ SD (100 simulation runs).}\label{fig:activations}
\end{figure}

So, we suspect that the one-hot method has larger linear-layer outputs compared to the population code. The larger values cause a problem for robustness because they make it more likely that a perturbation causes spurious activations that suppress the correct output value. To test this hypothesize, we evaluated the activations after the final linear layer, and compared one hot with population code. As a result, both the absolute max and min values were indeed larger for one hot for the 3, 5, and 8-layer networks (Fig. \ref{fig:activations}). For the 8-layer network, the values were about 5x larger.

\begin{figure}[h]
\centering
\includegraphics[width=0.49\textwidth]{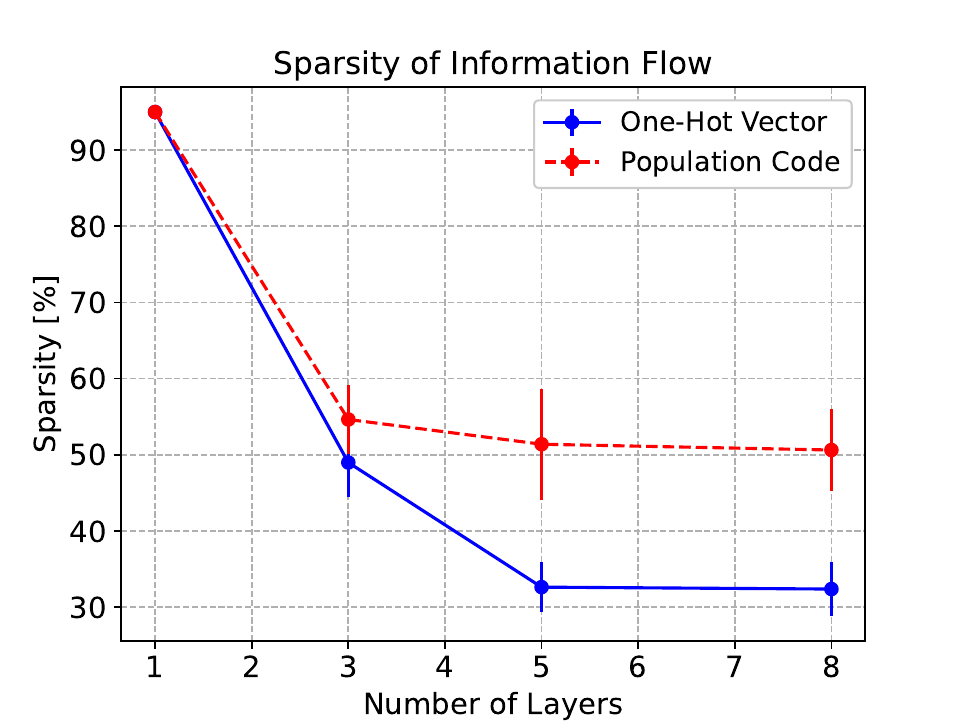}
\vspace{-5mm}
\caption{The population-code method showed a sparser information flow than the one-hot method. Values are mean $\pm$ SD (100 simulation runs).}\label{fig:sparsity}
\end{figure}

In addition, we observed that the information-flow through the network was sparser for the population-code method (Fig. \ref{fig:sparse_example}). Here, the flow through a connection is determined by multiplying the connection's weight by the activation level of the neuron sending the signal. To quantitatively evaluate the sparsity, we counted as unused those connections whose absolute-value flow was less than or equal to 1/30 of the maximum absolute flow in a given layer. Particularly, for 5 and 8-layer networks, the fraction of those connections below threshold was substantially smaller for the one-hot method compared to the population-code method; so, the latter had a sparser flow (Fig. \ref{fig:sparsity}). This sparsity may contribute the network's robustness to noise \cite{lecun1989,hassibi1992,wen2016,ahmad2019}.

\section{Experiments with Real-World Data}

We compared the population-code method against the single variable and one-hot methods for predicting object orientation from grayscale images. For this test, we used the T-LESS dataset of industry-relevant featureless objects \cite{hodan2017}. A particular challenge for this dataset is the ambiguity of object pose because most of the objects are symmetric and, therefore, have multiple equivalent pose representation. Here, we demonstrate that the population code can handle this ambiguity.

\subsection{Population Code for Object Orientation}

To encode the object orientation with a neural population code, we first transformed the rotation matrix into a rotation axis and a rotation angle. For the axis, we arranged the preferred values of each neuron on the surface of a sphere. To obtain a near uniform distribution of axes, we computed a spherical Fibonacci lattice \cite{dixon91}. For the rotation angle, the preferred values were arranged in a circle. To combine rotation axis with angle, our population code had for each point on the Fibonacci sphere a circle of neurons for the rotation angle. So, the space to represent the orientation is a direct product of a sphere and a circle. 

Given a rotation axis and angle, we activated all neurons using Gaussian tuning curves,

\begin{equation}
f_i = \exp\left(-\frac{d\theta_i^2 + d\phi_i^2}{2 \sigma^2}\right)\, ,
\end{equation}
where $d\theta_i$ is the angle between the encoded axis and the preferred axis on the sphere for neuron $i$, $d\phi_i$ the angle between the encoded angle and the preferred angle of neuron $i$, and $\sigma$ is the tuning width, here $20^\circ$.
 
\begin{figure}[h]
\centering
{\small Input Image\hspace{2.3cm}Population Code}\\ \vspace{2mm}
\includegraphics[width=0.48\textwidth]{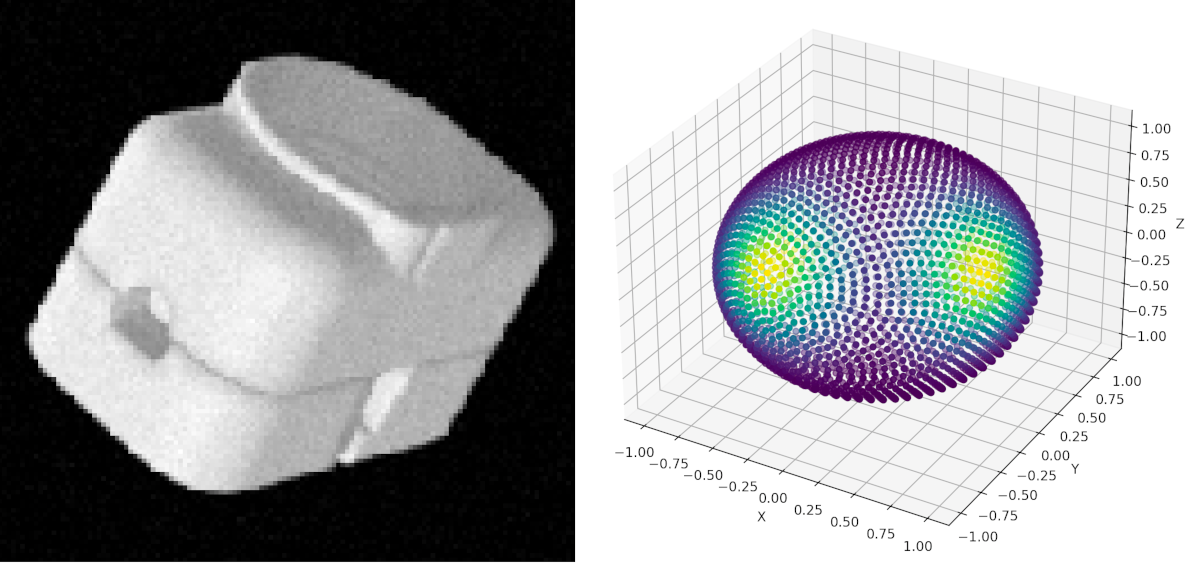}\\ \vspace{1mm}
\includegraphics[width=0.48\textwidth]{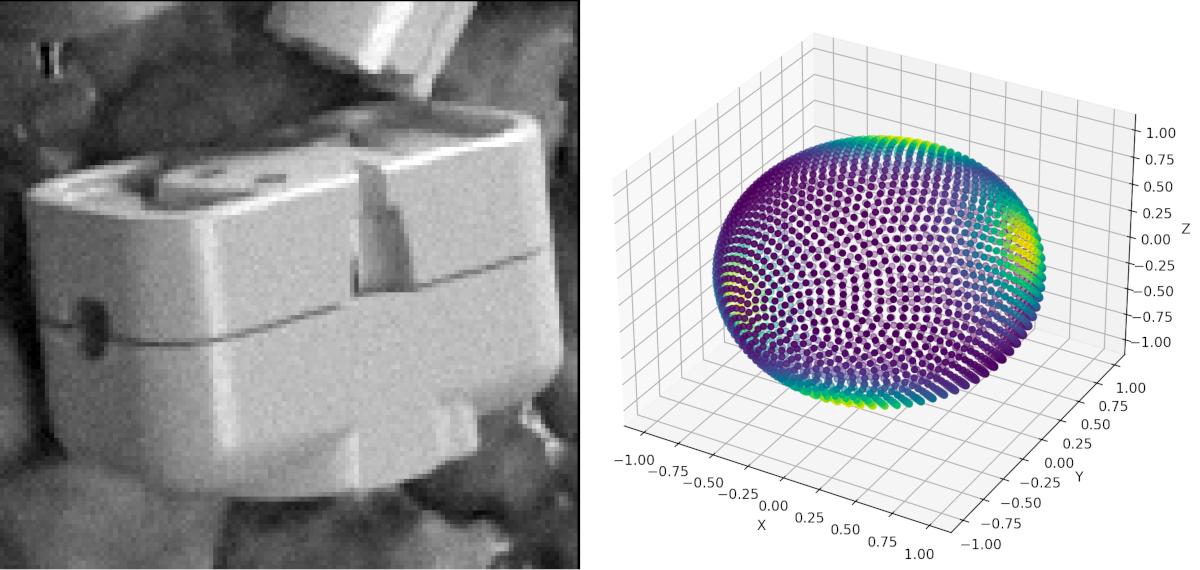}
\caption{The population code accounts for symmetry by having multiple peaks. Color shows neural activation (yellow: high; blue: low). Here, illustrating only the Fibonacci sphere for the axes.}\label{fig:code}
\end{figure}

 To compute the target population code from a ground truth transformation matrix, ${\bf R}_o$, we computed activations for all symmetry transformations of  ${\bf R}_o$,  
 \begin{equation}
 {\bf R'}_k = {\bf R}_o {\bf R}_{\mbox{\small sym}}^k\,,
 \end{equation}
where $\{ {\bf R}_{\mbox{\small sym}}^k\}$ is the set of symmetry transformations for a given object (for an application like manufacturing, the symmetry of an object is usually known a priori). Our target population code is the sum of all activations $f_i$ over all $k$ symmetry transformations (Fig. \ref{fig:code}).
 
 The transformation from rotation matrix to axis/angle, $\{{\bf r}, \phi\}$, is not unique since $\{-{\bf r}, 2 \pi - \phi\}$ is an equivalent axis/angle combination to $\{{\bf r}$, $\phi\}$. Therefore, for each symmetry transformation, we computed the tuning-curve activations also for $\{-{\bf r}, 2 \pi - \phi\}$ and added those to the target activations (resulting in four peaks in Fig. \ref{fig:code}).
 
For objects with a symmetry axis, which would result in infinitely many equivalent poses, we computed a variant of our population code: since the population code for the rotation angle would show uniform activations, we simplified the code and omitted the rotation angle neurons, using just the neurons encoding the rotation axis on the Fibonacci sphere. Here, the code was computed simply as
\begin{equation}
f_i = \exp\left(-\frac{d\theta_i^2 }{2 \sigma^2}\right)\, ,
\end{equation}
 without having to superimpose activations for symmetries. 
 
\subsection{Methods}

For predicting object orientation, we mapped gray-scale images of size 128x128 pixels onto the output layer. For the population code, we had a vector of size $n m$, where $n$ is the number of preferred axes and $m$ the number of preferred angles. Here, we used $n = 2,562$ and $m = 36$, matching the number of rotation candidates used in \cite{sundermeyer2020}. For the one-hot approach, we used a vector of the same size. 

For the single-variable approach, we directly mapped onto the 6-dimensions of the first two columns of the rotation matrix. Mapping onto two columns of {\bf R} has been shown to be advantageous over other representations like quaternions due to the continuity of the  {\bf R}6 space \cite{zhou2019}. This representation was also used by one of the best performing pose-estimation methods \cite{wang2021}. The entire rotation matrix can be reconstructed from the two columns using Gram-Schmidt orthonormalization. In the special case of objects with a symmetry axis, we directly mapped onto the 3 coordinates of the rotation axis.

\begin{figure*}[t]
      \centering
      \includegraphics[width=0.76\textwidth]{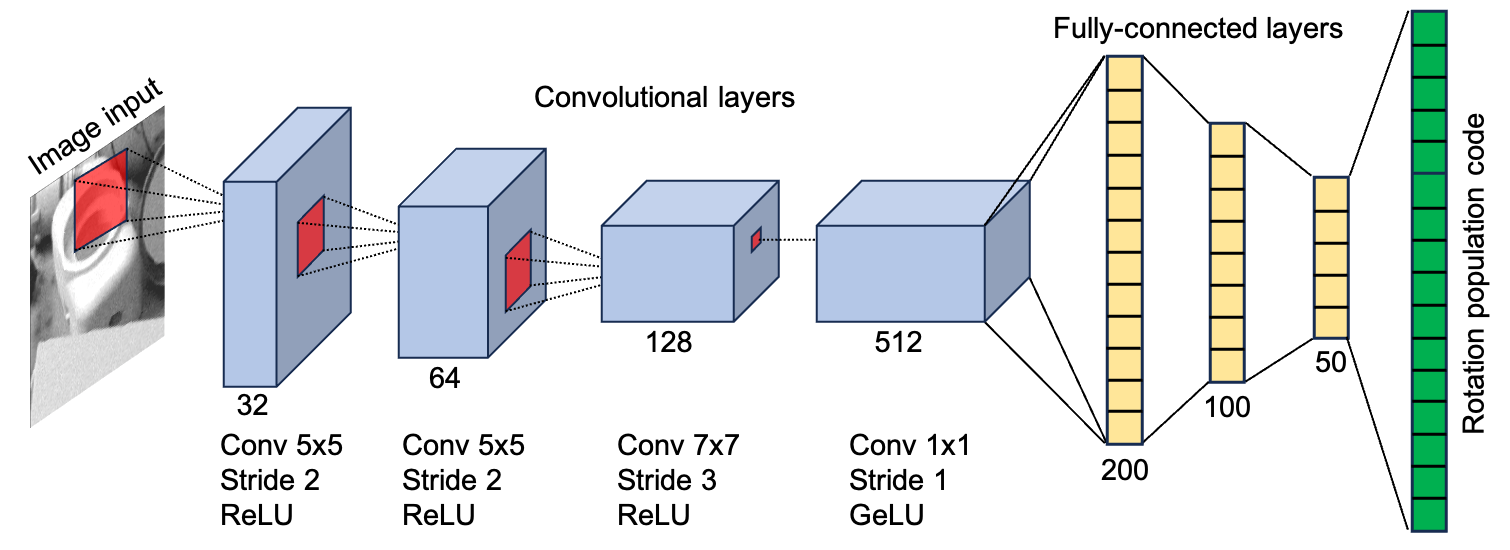}
      \vspace{-3mm}
      \caption{For the T-LESS experiment, our network architecture contained a sequence of 4 convolutional blocks and 4 linear layers}\label{fig:arch}
 \end{figure*}

The input images were fed through four convolutional layers (see Fig. \ref{fig:arch} regarding the kernel size and number of features). The first three convolutional layers were followed by batch normalization and ReLU activation functions. This architectural choice was inspired by the Augmented Autoencoder \cite{sundermeyer2020}. The fourth convolutional layer with kernel size 1x1 was followed by a GeLU non-linear function. This architectural element was inspired by \cite{liu2022}, which demonstrated its benefit.

After the convolutional layers, the features were mapped through three hidden linear layers onto the output linear layer. Each hidden layer was followed by a Leaky ReLU activation function. For the population code, we applied again a sigmoid function to the last linear layer. 

The loss function for the population code computed the MSE between the predicted and target code vector. For one-hot, we used again the cross-entropy loss. For the single-variable approach, we used the L1 norm in {\bf R}6 for discrete symmetry and the MSE on the rotation-axis coordinates for rotational symmetry. For discrete symmetries, the single-variable approach would simply fail because an input image did not have a unique output target. So, for multiple targets, we computed the minimum loss across the different equivalent target rotation matrices. Population code and one hot did not have this problem since the output vector can represent alternative targets.

For training, we used the images from the BlenderProc4BOP dataset \cite{hodavn2020}, extracting all instances of an object with at least 60\% visibility. We augmented these data with the T-LESS Primesense camera images \cite{hodan2017}. For all images, we used the provided bounding boxes and made square cutouts by using the maximum of width and height plus a 10\% padding. The resulting cutouts were scaled to 128x128 pixels. We supplemented these images with 8,000 generated images per object using pyrender and the 3D model files from the T-LESS dataset. This combination resulted in about 20,000 to 22,000 images per object.

\begin{figure}[h]
\centering
\includegraphics[width=0.11\textwidth]{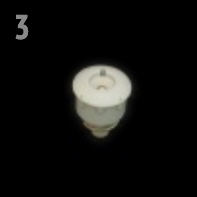}
\includegraphics[width=0.11\textwidth]{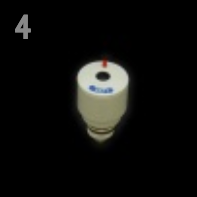}
\includegraphics[width=0.11\textwidth]{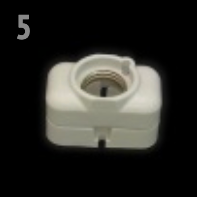}
\includegraphics[width=0.11\textwidth]{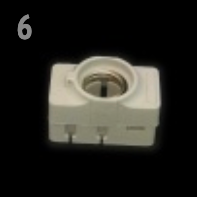}
\caption{Selected objects from the T-LESS dataset}\label{fig:objects}
\end{figure}

To further increase the amount of training data, we combined some pairs visually similar objects into one dataset and trained a single model for both objects. Object 4 was training on objects 3 and 4, and object 5 was trained on objects 5 and 6 (Fig. \ref{fig:objects}).

For training, we used the Adam optimizer with a learning rate of 0.0002 and a batch size of 64. We trained all methods for 80 epochs. During training, we augmented the data for the pyrender-generated images by 1) adding random brightness, a uniformly-distributed value between -0.2 and 0.2 to all pixels with 50\% probability and 2) adding a background image with 70\% probability. Moreover, for all training images, we added Gaussian noise (SD: 0.02) with 50\% probability, did contrast normalization, and shifted the image in the image plane ($\pm 5$ pixels in x and y) with 90\% probability.
 
\subsection{Experimental Results}

For evaluation, we computed the three metrics used by the BOP: Benchmark for 6D Object Pose Estimation \cite{hodan2020}: Visible Surface Discrepancy (VSD), Maximum Symmetry-Aware Surface Distance (MSSD), and Maximum Symmetry-Aware Projection Distance (MSPD). The corresponding accuracy measures are defined in \cite{hodan2020}. In addition, we used the official T-LESS test set for the BOP Pose Estimation Challenge \cite{hodavn2020}. 

Among all methods, population code had the highest accuracies (Tab. \ref{tab:obj4} and \ref{tab:obj5}). We carried out the comparison between methods only for objects 4 and 5 because the performance differences were already quite large. For completion, we computed the results for all 30 T-LESS objects for the population code. 
The average MSSD accuracy was 84.92\%,  MSPD 84.09\%, and VSD 74.06\%. These results are competitive considering only grayscale image input \cite{bop24}.
  
\begin{table}[h]
\caption{Comparison of methods for object 4 in the T-LESS dataset. For the single-variable approach, we predicted the rotation axis.}\label{tab:obj4}
\vspace{-1mm}
\centering
\resizebox{0.4\textwidth}{!}{
\begin{tabular}{|l||c|c|c|c|c|c|}
    \hline
    Approach / Metric         & VSD    & MSSD  & MSPD   \\ \hline \hline
    Population Code  & \bf{75.8\%} & \bf{86.3\%} & \bf{88.4\%}  \\ \hline
    One-Hot Vector    & 55.0\% & 63.7\% & 67.5\%  \\ \hline
    Single Variable          & 18.9\% & 23.7\% & 26.4\% \\ \hline
    Random Weights  & 8.2\% & 9.0\% & 14.4\%   \\ \hline   
    \end{tabular}
}
\end{table}
\vspace{-3mm}
 \begin{table}[h]
\caption{Comparison of methods for object 5 in the T-LESS dataset. For the single-variable approach, we used multiple targets to deal with the pose ambiguity.}\label{tab:obj5}
\vspace{-1mm}
\centering
\resizebox{0.4\textwidth}{!}{
\begin{tabular}{|l||c|c|c|c|c|c|}
    \hline
    Approach / Metric       & VSD    & MSSD  & MSPD   \\ \hline \hline
    Population Code  & \bf{83.6\%} & \bf{89.4\%} & \bf{87.4\%} \\ \hline
    One-Hot Vector    & 63.4\% & 63.8\% & 61.1\%   \\ \hline
    Single Variable    & 38.7\% & 22.9\% & 18.7\% \\ \hline
    Random Weights & 17.4\% & 1.7\% & 0.9\%     \\ \hline   
    \end{tabular}
}
\end{table}
 
\section{Conclusions}

We found that using a population code as the output of a deep network, rather than directly mapping onto prediction variables, offers the following advantages:

\begin{itemize}
\item Increased robustness to input noise,
\item Higher accuracies in real-world prediction tasks, and
\item Ability to handle ambiguous output by effectively representing a multimodal probability distribution.
\end{itemize}

Moreover, a population code has better noise robustness and prediction accuracy compared to a one-hot vector of the same size. The reason for this advantage might be the sparser information flow that we found with population codes and the reduction of extreme output values in the last linear layer of a trained network.

For many practical applications, the population-code output still has to be decoded, e.g., to read out an object's pose. Here, for simplicity, we just used the preferred value of the neuron with the maximum activation. This strategy is essentially the same as decoding a one-hot vector. But given the research on decoding population codes, many alternative methods already exist \cite{seung1993,salinas1994,baldi1988, pouget1998}. So, our results could be further improved. For example, using the maximum likelihood estimate based on the probability distribution of a target variable would likely not only improve noise robustness but also enable decoding of variables at a finer resolution than permitted by population-code sampling. Such an approach would allow for a sparser sampling, which would benefit prediction tasks with higher-dimensional outputs.

In our future research, we will apply population codes to other prediction tasks and have already observed improvements in various domains.

{
    \small
    \bibliographystyle{ieeenat_fullname}
    \bibliography{main}
}

% WARNING: do not forget to delete the supplementary pages from your submission 
% \input{sec/X_suppl}

\end{document}